# MODULARITY IN A CONNECTIONIST MODEL OF MORPHOLOGY ACQUISITION

Michael Gasser

Departments of Computer Science and Linguistics
Indiana University

## Abstract

This paper describes a modular connectionist model of the acquisition of receptive inflectional morphology. The model takes inputs in the form of phones one at a time and outputs the associated roots and inflections. In its simplest version, the network consists of separate simple recurrent subnetworks for root and inflection identification; both networks take the phone sequence as inputs. It is shown that the performance of the two separate modular networks is superior to a single network responsible for both root and inflection identification. In a more elaborate version of the model, the network learns to use separate hidden-layer modules to solve the separate tasks of root and inflection identification.

## INTRODUCTION

For many natural languages, the complexity of bound morphology makes it a potentially challenging problem for a learning system, whether human or machine. A language learner must acquire both the ability to map polymorphemic words onto the sets of semantic elements they represent and to map meanings onto polymorphemic words. Unlike previous work on connectionist morphology (e.g., MacWhinney & Leinbach (1991), Plunkett & Marchman (1991) and Rumelhart & McClelland (1986)), the focus of this paper is receptive morphology, which represents the more fundamental, or at least the earlier, process, one which productive morphology presumably builds on.

The task of learning receptive morphology is viewed here as follows. The learner is "trained" on pairs of forms, consisting of sequences of phones, and "meanings", consisting of sets of roots and inflections. I will refer to the task as root and inflection *identification*. Generalization is tested by presenting the learner with words consisting of novel combinations of familiar morphemes. If the rule in question has been acquired, the learner is able to identify the root and inflections in the test word.

Of interest is whether a model is capable of acquiring rules of all of the types known for natural languages. This paper describes a psychologically motivated connectionist model (Modular Connectionist Network for the Acquisition of Morphology, MCNAM) which approaches this level of performance. The emphasis here is on the role of modularity at the level of root and inflection in the model. I show how this sort of modularity improves performance dramatically and consider how a network might learn to use modules it is provided with. A separate paper (Gasser, 1994) looks in detail at the model's performance for particular categories of morphology, in particular, template morphology and reduplication.

The paper is organized as follows. I first provide a brief overview of the categories of morphological rules found in the world's languages. I then present a simple version of the model and discuss simulations which demonstrate that it generalizes for most kinds of morphological rules. I then describe a version of the model augmented with modularity at the level of root and inflection which generalizes significantly better and show why this appears to be the case. Finally, I describe some tentative attempts to develop a model which is provided with modules and **learns** how to use them to solve the morphology identification tasks it is faced with.

## CATEGORIES OF MORPHOLOGICAL PROCESSES

I will be discussing morphology in terms of the traditional categories of "root" and "inflection" and morphological processes in terms of "rules", though it should be emphasized that a language learner does not have direct access to these notions, and it is an open question whether they need to be an explicit part of the system which the learner develops, let alone the device which the learner starts out with. I will not make a distinction between inflectional and derivational morphology (using "inflection" for both) and will not consider compounding.

*Affixation* involves the addition of the inflection to the root (or stem), either before (*prefixation*), after (*suffixation*), within (*infixation*), or both before and after (*circumfixation*) the root. A further type of morphological rule, which I will refer to as *mutation*, consists in modification to the root segments themselves. A third type of rule, familiar in Semitic languages, is known as *template* morphology. Here a word (or stem) consists of a root and a pattern of segments which are intercalated between the root segments in a way which is specified within the pattern. A fourth type, the rarest of all, consists in the *deletion* of one or more segments. A fifth type, like affixation, involves the addition of something to the root form. But the form of what is added in this case is a copy, or a systematically altered copy, of some portion of the root. This process, *reduplication*, is in one way the most complex type of morphology (though it may not necessarily be the most difficult for a child to learn) because it seems to require a variable. It is not handled by the model discussed in this paper. Gasser (1994) discusses modification of the model which is required to accommodate reduplication.

## THE MODEL

The approach to language acquisition exemplified in this paper differs from traditional symbolic approaches in that the focus is on specifying the sort of cognitive *architecture* and the sort of general processing and learning *mechanisms* which have the capacity to learn some aspect of language, rather than the innate *knowledge* which this might require. If successful, such a model would provide a simpler account of the acquisition of morphology than one which begins with symbolic knowledge and constraints. Connectionist models are interesting in this regard because of their powerful sub-symbolic learning algorithms. But in the past, there has been relatively little interest in investigating the effect on the language acquisition capacity of structuring networks in particular ways. The concern in this paper will be with what is gained by adding modularity to a network.

Given the basic problem of what it means to learn receptive morphology, I will begin with one of the simplest networks that could have that capacity and then augment the device as necessary. In this paper, two versions of the model are described. Version 1 successfully learns simple examples of all of the morphological rules except reduplication and circumfixation, but its performance is far from the level that might be expected from a human language learner. Version 2 (MCNAM proper) incorporates a form of built-in modularity which separates portions of the network responsible for the identification of the root and the inflections; this improves the network's performance significantly on all of the rule types except reduplication, which cannot be learned even by a network outfitted with this form of modularity.

Word recognition is an incremental process. Words are often recognized long before they finish; hearers seem to be continuously comparing the contents of a linguistic short-term memory with the phonological representations in their mental lexicons (Marslen-Wilson & Tyler, 1980). Thus the task at hand requires a short-term memory of some sort. There are several ways of representing short-term memory in connectionist networks (Port, 1990), in particular, through the use of time-delay connections out of input units and through the use of recurrent time-delay connections on some of the network units. The most flexible approach makes use of recurrent connections on hidden units, though the arguments in favor of this option are beyond the scope of this paper. The model to be described here is a network of this type, a version of the simple recurrent network due to Elman (1990).

### Version 1

The Version 1 network is shown in Figure 1. Each box represents a layer of connectionist processing units and each arrow a complete set of weighted connections between two layers. The network operates as follows. A sequence of phones is presented to the input layer one at a time. That is, each tick of the network's clock represents the presentation of a single phone. Each phone unit represents a phonetic feature, and each word consists of a sequence of phones preceded by a boundary "phone" made up of 0.0 activations.

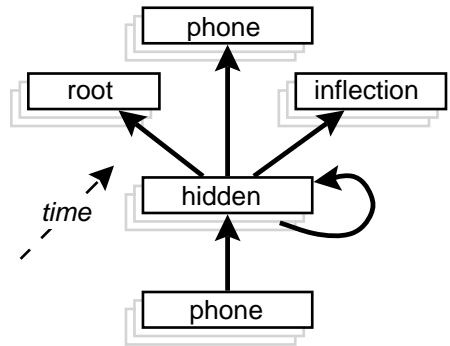

Figure 1: Network for Acquisition of Morphology (Version 1)

An input phone pattern sends activation to the network's hidden layer. The hidden layer also receives activation from the pattern that appeared there on the previous time step. Thus each hidden unit is joined by a time-delay connection to each other hidden unit. It is the previous hidden-layer pattern which represents the system's short-term memory. Because the hidden layer has access to this previous state, which in turn depended on its state at the time step before that, there is no absolute limit to the length of the context stored in the short-term memory. At the beginning of each word sequence, the hidden layer is reinitialized to a pattern consisting of 0.0 activations.

Finally the output units are activated by the hidden layer. There are three output layers. One represents simply a copy of the current input phone. Training the network to auto-associate its current input aids in learning the root and inflection identification task because it forces the network to learn to distinguish the individual phones at the hidden layer, a prerequisite to using the short-term memory effectively. The second layer of output units represents the root "meaning". For each root there is a single output unit. Thus while there is no real semantics, the association between the input phone sequence and the "meaning" is at least an arbitrary one. The third group of output units represents the inflection "meaning". Again there is a unit for each separate inflection.

For each input phone, the network receives a target consisting of the correct phone, root, and inflection outputs for the current word. The phone target is identical to the input phone. The root and inflection targets, which are constant throughout the presentation of a word, are the patterns associated with the root and inflection for the input word.

The network is trained using the backpropagation learning algorithm (Rumelhart, Hinton, & Williams, 1986), which adjusts the weights on all of the network's connections in such a way as to minimize the error, that is, the difference between the network's outputs and the targets. For each morphological rule, a separate network is trained on a subset of the possible combinations of root and inflection. At various points during training, the network is tested on unfamiliar words, that is, novel combinations of roots and inflections. The performance of the network is the percentage of the test roots and inflections for which its output is correct at the end of each word sequence when it has enough information to identify both root and inflection. A "correct" output is one which is closer to the appropriate target than to any of the others.

In all of the experiments reported on here, the stimuli presented to the network consisted of words in an artificial language. The phoneme inventory of the language was made up 19 phones (24 for the mutation rule, which nasalizes vowels). For each morphological rule, there were 30 roots, 15 each of CVC and CVCVC patterns of phones. Each word consisted of two morphemes, a root and a single "tense" inflection, marking the "present" or "past". Examples of each rule: (1) suffix: present–*vibuni*, past–*vibuna*; (2) prefix: present–*ivibun*, past–*avibun*; (3) infix: present–*vikbun*, past–*vinbun*; (4) circumfix: present–*ivibuni*, past–*avibuna*; (5) mutation: present–*vibun*, past–*vibũn*; (6) deletion: present–*vibun*, past–*vibu*; (7) template: present–*vaban*, past–*vbaan*.

For each morphological rule there were 60 (30 roots × 2 inflections) different words. From these 40 were selected randomly as training words, and the remaining 20 were set aside as test words. For each rule, ten separate networks, with different random initial weights, were trained for 150 epochs (repetitions of all training patterns). Every 25 epochs, the performance of the network on the test patterns was assessed.

Figure 2 shows the performance of the Version 1 network on each rule (as well as performance on Version 2, to be described below). Note that chance performance for the roots was .033 and for the inflections .5 since there were 30 roots and 2 inflections. There are several things to notice in these results. Except for root identification for the circumfix rule, the network performs well above chance. However, the results are still disappointing in many cases. In particular, note the poor performance on root identification for the prefix rule and inflection identification for the suffix rule. The behavior is much poorer than we might expect from a child learning these relatively simple rules.

The problem, it turns out, is interference between the two tasks which the network is faced with. On the one hand, it must pay attention to information which is relevant to root identification, on the other, to information relevant to inflection identification. This means making use of the network's short-term memory in very different ways. Consider the prefixing case, for example. Here for inflection identification, the network need only pay attention to the first phone and then remember it until the end of the sequence is reached, ignoring all of the phones which appear in between. For root identification, however, the network does best if it ignores the initial phone in the sequence and then pays careful attention to each of the following phones.

Ideally the network's hidden layer would divide into modules, one dedicated to root identification, the other to inflection identification. This could happen if some of the recurrent hidden-unit weights and some of the weights on hidden-to-output connections went to 0. However, ordinary backpropagation tends to implement sharing among hidden-layer units: each hidden-layer unit participates to some extent in activating all output units. When there are conflicting output tasks, as in this case, there are two sorts of possible consequences: either performance on both tasks is mediocre, or the simpler task comes to dominate the hidden layer, yielding good performance on that task and poor performance on the other. In the Version 1 results shown in Figure 2, we see both sorts of outcomes.

What is apparently needed is modularity at the hidden-layer level. One sort of modularity is hard-wired into the network's architecture in Version 2 of the model, described in the next section.

**Version 2**

Because root and inflection identification make conflicting demands on the network's short-term memory, it is predicted that performance will improve with separate hidden layers for the two tasks. Various degrees of modularity are possible in connectionist networks; the form implemented in Version 2 of the model is total modularity, completely separate networks for the two tasks. This is shown in Figure 3. There are now two hidden-layer modules, each with recurrent connections only to units within the same module and with connections to one of the two output identification layers of units. (Both hidden layers connect to the auto-associative phone output layer.)

The same stimuli were used in training and test-ing the Version 2 network as the Version 1 network. Each Version 2 network had the same number of total hidden units as each Version 1 network, 30. Each hidden-layer module contained 15 units. Note that this means there are fewer connections in the Version 2 than the Version 1 networks. Investigations with networks with hidden layers of different sizes indicate that, if anything, this should favor the Version 1 networks.

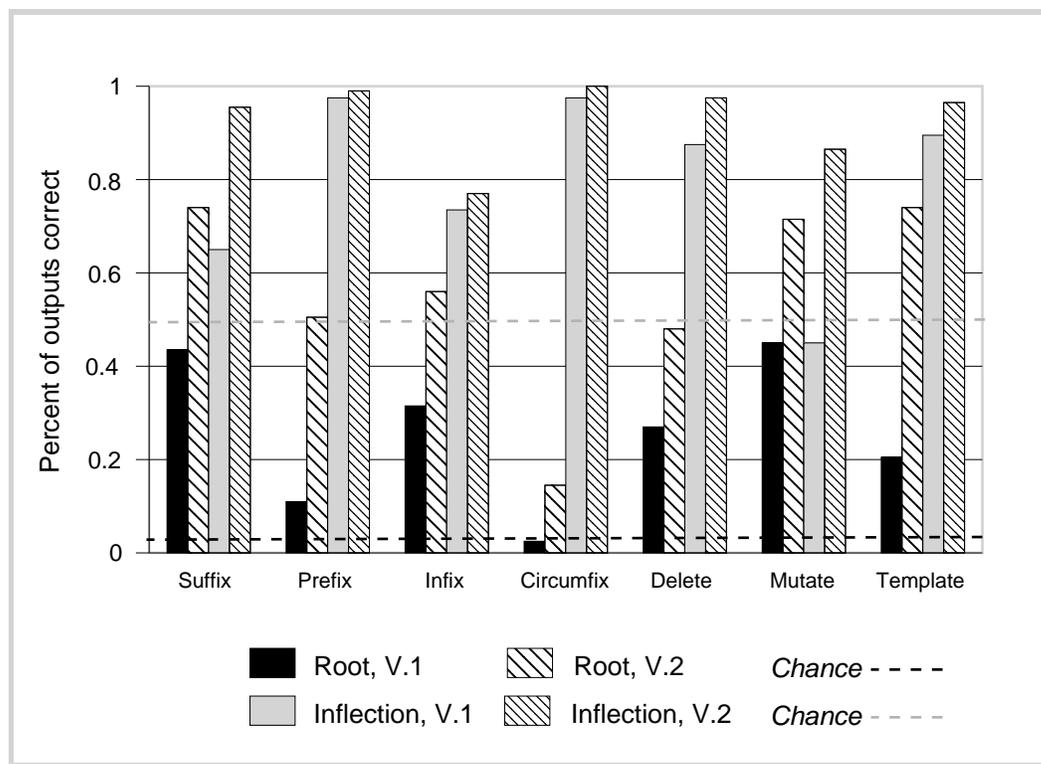

Figure 2: Performance on Test Words Following Training (Network Versions 1 and 2)

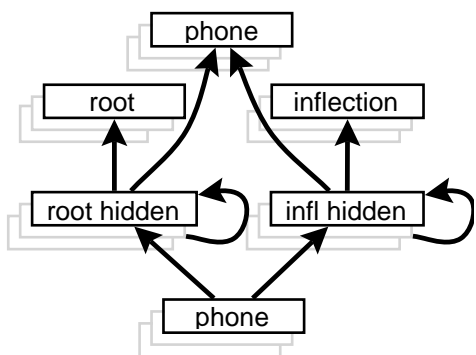

Figure 3: Network for Acquisition of Morphology (Version 2)

Figure 2 compares results from the two versions following 150 epochs of training. For all of the rule types, modularity improves performance for both root and inflection identification. Obviously, hidden-layer modularity results in diminished interference between the two output tasks. Performance is still far from perfect for some of the rule types, but further improvement is possible with optimization of the learning parameters.

## TOWARDS ADAPTIVE MODULARITY

It is important to be clear on the nature of the modularity being proposed here. As discussed above, I have defined the task of word recognition in such a way that there is a built-in distinction between lexical and grammatical "meanings" because these are localized in separate output layers. The modular architecture of Figure 3 extends this distinction into the domain of phonology. That is, the shape of words is represented internally (on the hidden layer) in terms of two distinct patterns, one for the root and one for the inflection, and the network "knows" this even before it is trained, though of course it does not know how the root and inflections will be realized in the language.

A further concern arises when we consider what happens when more than one grammatical category is represented in the words being recognized, for example, aspect in addition to tense on verbs. Assuming the hidden-layer modules are a part of the innate makeup of the learning device, this means that a fixed number of given modules must be divided up among the separate output "tasks" which the target language presents. Ideally, the network would have the capacity to figure out for itself how to distribute the modules it starts with among the various output tasks; I return to this possibility below. But it is also informative to investigate what sort of a sharing arrangement achieves the best performance. For example, given two modules and three output tasks, root identification and the identification of two separate inflections, which of the three possible ways of sharing the modules achieves the best performance?

Two sets of experiments were conducted to investigate the optimal use of fixed modules by a network, one designed to determine the best way of distributing modules among output tasks when the number of modules does not match the number of output tasks and one designed to determine whether a network could assign the modules to the tasks itself. In both sets of experiments, the stimuli were words composed of a stem and two affixes, either two suffixes, two prefixes, or one prefix and one suffix. (All of these possibilities occur in natural languages.) The roots were the same ones used in the affixation and deletion experiments already reported. In the two-suffix case, the first suffix was /a/ or /i/, the second suffix /s/ or /k/. Thus the four forms for the root *migon* were *migonik*, *migonis*, *migonak*, and *migonas*. In the two-prefix case the prefixes were /s/ or /k/ and /a/ or /i/. In the prefix–suffix case, the prefix was /u/ or /e/ and the suffix /a/ or /i/. There were in all cases two hidden-layer modules. The size of the modules was such that the root identification task had potentially 20 units and each of the inflection identification tasks potentially 3 units at its disposal; the sum of the units in the two modules was always 26.

The results are only summarized here. The configuration in which a single module is shared by the two affix-identification tasks is consistently superior for peformance on root identification but only superior for affix identification in the two-suffix case. For the prefix-suffix case, the configuration in which one module is shared by root identification and suffix identification is clearly inferior to the other two configurations for performance on suffix identification. For the two-prefix case, the configurations make little difference for performance on identification of either of the prefixes. Note that the results for the two-prefix and two-suffix cases agree with those for the single-prefix and single-suffix cases respectively (Figure 2).

What the results for root identification make clear is that, even though the affix identification tasks are easily learned with only 3 units, when they are provided with more units (23 in these experiments), they will tend to "distribute" themselves over the available units. If this were not the case, performance on the competing, and more difficult, task, root identification, would be no better when it has 20 units to itself than when it shares 23 units with one of the other two tasks.

We conclude that the division of labor into separate root and inflection identification modules works best, primarily because it reduces interference with root identification, but also for the two-suffix case, and to a lesser extent for the prefix-suffix case, because it improves performance on affix identification. If one distribution of the available modules is more efficient than the others, we would like the network to be able to find this distribution on its own. Otherwise it would have to be wired into the system from the start, and this would require knowing that the different inflection tasks belong to the same category. Some form of *adaptive* use of the available modules seems called for.

Given a system with a fixed set of modules but no wired-in constraints on how they are used to solve the various output tasks, can a network organize itself in such a way that it uses the modules efficiently? There has been considerable interest in the last few years in architectures which are endowed with modularity and learn to use the modularity to solve tasks which call for it. The architecture described by Jacobs, Jordan, & Barto (1991) is an example. In this approach there are connections from each modular hidden layer to all of the output units. In addition there are one or more gating networks whose function is to modulate the input to the output units from the hidden-layer modules. In the version of the architecture which is appropriate for domains such as the current one, there is a single gating unit responsible for the set of connections from each hidden module to each output task group. The outputs of the modules are weighted by the outputs of the corresponding gating units to give the output of the entire system. The whole network is trained using backpropagation. For each of the modules, the error is weighted by the value of the gating input as it is passed back to the modules. Thus each module adjusts its weights in such a way that the difference between the system's output and the desired target is minimized, and the extent to which a module's weights are changed depends on its contribution to the output. For the gating networks, the error function implements competition among the modules for each output task group. For our purposes, two further augmentations are required. First, we are dealing with recurrent networks, so we permit each of the modular hidden layers to see its own previous values in addition to the current input, but not the previous values of the hidden layers of the other modules. Second, we are interested not only in competition among the modules for the output groups, but also in competition among the output groups for the modules. In particular, we would like to prevent the network from assigning a single module to all output tasks.

To achieve this, the error function is modified so that error is minimized, all else being equal, when the total of the outputs of all gating units dedicated to a single module is neither close to 0.0 nor close to the total number of output groups.

Figure 4 shows the architecture for the situation in which there is only one inflection to be learned. (The auto-associative phone output layer is not shown.) The connections ending in circles symbolize the competition between sets of gating units which is built into the error function for the network. Note that the gating units have no input connections. These units have only to learn a bias, which, once the system is stable, leads to a relatively constant output. The assumption is that, since we are dealing with a spatial crosstalk problem, the way in which particular modules are assigned to particular tasks should not vary with the input to the network.

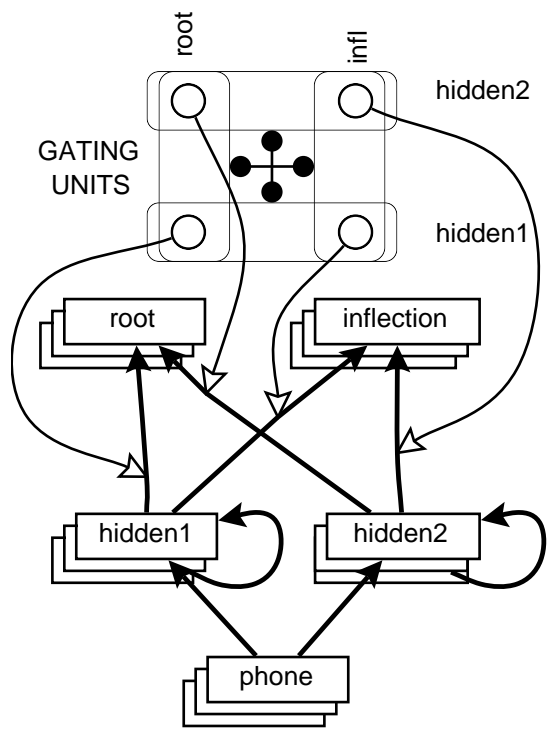

Figure 4: Adaptive Modular Architecture for Morphology Acquisition

An initial experiment demonstrated that the adaptive modular network consistently assigned separate modules to the output tasks when there were two modules and two tasks (identification of the root and a single inflection).

Next a set of experiments tested whether the adaptive modular architecture would assign two modules to three tasks (root and two inflections) in the most efficient way for the two-suffix, two-prefix, and prefix-suffix cases. Recall that the most efficient pattern of connectivity in all cases was the one in which one of the two modules was shared by the two affix identification tasks.

Adaptive modular networks with two modules of 15 units each were trained on the two-suffix, two-prefix, and prefix-suffix tasks described in the last section. Following 120 epochs, the outputs of the six gating units for the different modules were examined to determine how the modules were shared. The results were completely negative; the three possible ways of assigning the modules to the three identification tasks occurred with approximately equal frequency. The problem was that the inflection identification tasks were so much easier than the root identification task that they claimed the two modules for themselves early on, while neither module was strongly preferred by the root task. Thus as often as not, the two inflections ended up assigned to different modules. To compensate for this, then, is it reasonable to give root identification some sort of advantage over inflection identification? It is well-known that children begin to acquire lexical morphemes before they acquire grammatical morphemes. Among the reasons for this is probably the more abstract nature of the meanings of the grammatical morphemes. In terms of the network's tasks, this relative difficulty would translate into an inability to know what the inflection targets would be for particular input patterns. Thus we could model it by delaying training on the inflection identification task.

The experiment with the adaptive modular networks was repeated, this time with the following training regimen. Entire words (consisting of root and two affixes) were presented throughout training, but for the first 80 epochs, the network saw targets for only the root identification task. That is, the connections into the output units for the two inflections were not altered during this phase. Following the 80th epoch, by which time the network was well on its way to learning the roots, training on the inflections was introduced. This procedure was followed for the two-suffix, two-prefix, and prefix-suffix tasks; 20 separate networks were trained for each type. For the two-suffix task, in all cases the network organized itself in the predicted way. That is, for all 20 networks one of the modules was associated mainly with the two inflection output units and the other associated with the root output units. In the prefix-suffix case, however, the results were more equivocal. Only 12 out of 20 of the networks organized themselves in such a way that the two inflection tasks were shared by one module, while in the 8 other cases, one module was shared by the root and prefix identification tasks. Finally, in the two-prefix case, all of the networks organized themselves in such a way that the root and the first prefix shared a module rather than in the apparently more efficient configuration.

The difference is not surprising when we consider the nature of the advantage of the configuration in which the two inflection identification tasks are shared by one module. For all three types of affixes, roots are identified better with this configuration. But this will have little effect on the way the network organizes itself because, following the 80th epoch when competition among the three output tasks is introduced, one or the other of the modules will already be firmly linked to the root output layer. At this point, the outcome will depend mainly on the competition between the two inflection identification tasks for the two modules, the one already claimed for root identification and the one which is still unused. Thus we can expect this training regimen to settle on the best configuration only when it makes a significant difference for inflection, as opposed to root, identification. Since this difference was greater for the two-suffix words than for the prefix-suffix words and virtually non-existent for the two-prefix words, there is the greatest preference in the two-suffix case for the configuration in which the two inflection tasks are shared by a single module. It is also of interest that for the prefix-suffix case, the network never chose to share one module between the root and the suffix; this is easily the least efficient of the three configurations from the perspective of inflection identification.

Thus we are left with only a partial solution to the problem of how the modular architecture might arise in the first place. For circumstances in which the different sorts of modularity impinge on inflection identification, the adaptive approach can find the right configuration. When it is performance on root identification that makes the difference, however, this approach has nothing to offer. Future work will also have to address what happens when there are more than two modules and/or more than two inflections in a word.

## CONCLUSIONS

Early work applying connectionist networks to high-level cognitive tasks often seemed based on the assumption that a single network would be able to handle a wide range of phenomena. Increasingly, however, the emphasis is moving in the direction of special-purpose modules for subtasks which may conflict with each other if handled by the same hardware (Jacobs et al., 1991). These approaches bring connectionist models somewhat more in line with the symbolic models which they seek to replace. In this paper I have shown how the ability of simple recurrent networks to extract "structure in time" (Elman, 1990) is enhanced by built-in modularity which permits the recurrent hidden-unit connections to develop in ways which are suitable for the root and inflection identification tasks. Note that this modularity does not amount to endowing the network with the distinction between root and affix because both modules take the entire sequence of phones as input, and the modularity is the same when the rule being learned is one for which there are no affixes at all (mutation, for example).

Modular approaches, whether symbolic or connectionist, inevitably raise further questions, however. The modularity in the pre-wired version of MCNAM, which is reminiscent of the traditional separation of lexical and grammatical knowledge in linguistic models, assumes that the division of "semantic" output units into lexical and grammatical categories has already been made. The adaptive version partially addresses this shortcoming, but it is only effective in cases where modularity benefits inflection identification. Furthermore, it is still based on the assumption that the output is divided initially into groups representing separate competing tasks. I am currently experimenting with related adaptive approaches, as well as methods involving weight decay and weight pruning, which treat each output *unit* as a separate task.